# Unsupervised Anomaly Detection Using Diffusion Trend Analysis for Display Inspection

*Eunwoo Kim*, Un Yang*, Cheol Lae Roh*, and Stefano Ermon***

***Mechatronics Technology Research Center, Samsung Display, Yongin-si, Gyeonggi-do Korea**

****Computer Science Department, Stanford University, Stanford, CA USA**

## Abstract

***Reconstruction-based anomaly detection via denoising diffusion model has limitations in determining appropriate noise parameters that can degrade anomalies while preserving normal characteristics. Also, normal regions can fluctuate considerably during reconstruction, resulting in false detection. In this paper, we propose a method to detect anomalies by analysis of reconstruction trend depending on the degree of degradation, effectively solving the both problems that impede practical application in display inspection.***

## 1. Introduction

Unsupervised anomaly detection is widely used in industrial inspection to build a line of defense against unexpected defects, and reconstruction-based approaches are particularly useful because they can detect both the anomaly's location and shape [1, 2]. In principle, the reconstruction generates "normal" image that acts as a standard to be compared with anomaly. An image is reconstructed based on a model, and if the reconstruction obtained is different from the original image, an anomaly is likely. As the generation of a proper normal image is essential to detect anomalies, many recent studies have adopted diffusion models due to their ability to produce high-quality normal images [3-5]. Specifically, diffusion-based anomaly detection first degrades the input image to some degree by adding noise, and then reconstructs the degraded image using a diffusion model trained only on normal images [6, 7]. The diffusion model denoises noisy images to follow the distribution of normal data, so if the input anomaly image is noisy enough to be indistinguishable from normal, the reconstructed image can be considered normal. Therefore, the reconstructed normal image can be compared to the original anomaly image to obtain an anomaly score, based on the assumption that the reconstruction error in anomaly regions is larger than the error in normal regions.

While diffusion-reconstruction-based anomaly detection has advantages, it also has two major drawbacks. First, it is difficult to determine the proper noise level to degrade the input image since larger noise levels reduce anomalies, but they can also distort normal regions. This challenge is particularly acute in display manufacturing, where inspection processes must contend with a diverse array of products, multiple production stages, and a wide spectrum of defect types. Moreover, this complexity impedes rapid adaptation to new products, necessitating time-consuming recalibration and validation. Second, diffusion models can create large variations even in normal regions during reconstruction, resulting in false positives. This is due to the high diversity of the diffusion model, which generates samples covering a wide range of normal distributions [8]. The high-volume nature of display manufacturing magnifies the risk of false detection, resulting in substantial resource inefficiencies.

In this work, we propose a novel approach that addresses both issues by focusing on the trend of reconstruction with increasing degradation noise instead of relying solely on reconstruction errors at a single noise level. When an anomaly image is reconstructed with a diffusion model after noise is added, the resulting reconstruction has a gradual trend as the noise added increases. Since the denoising diffusion model is trained to generate a normal image, the gradual degradation of the anomaly by noise added creates a trend in which it becomes increasingly normal. This trend is independent of the morphological features of the anomaly, which solves the problem of determining the proper noise level in the conventional method. The trend-based approach also reduces false positives in normal regions because it distinguishes reconstructed normal variants from anomaly regions, based on the observation that the fluctuation in normal regions does not have a directional trend with noise level, whereas anomaly regions has a change in one direction by gradually progressing from anomaly to normal.

We demonstrated that the proposed method outperformed the baseline, achieving 8.0% higher mAUROC and 20.3% higher mAP on public industrial anomaly detection datasets, while improving mAUROC by 10.0% and mAP by 9.4% on proprietary display inspection datasets.

## 2. Background

A denoising diffusion model is a type of generative model that learns to generate data by reversing a diffusion process. It involves a forward diffusion process that maps data to noise and a learned reverse process for data generation. These models have shown remarkable results in high-fidelity image generation, outperforming other generative models like generative adversarial networks (GANs) [8], and have applications in image editing, super resolution, and more due to their ability to provide high mode coverage of the training data distribution [3-5].

Diffusion model is useful in anomaly detection due to the ease of generating high-quality anomaly-free images from anomaly images, even though the diffusion model was trained only on normal images. Reconstruction-based anomaly detection using diffusion models works by adding noise and reconstructing it to generate a normal image, which is then compared to the original anomaly image. If the diffusion model replaces the anomaly with normal while preserving the original information, only the anomaly region is different, allowing the reconstruction error to become the anomaly score. Based on this principle, diffusion-based anomaly detection methods have been recently proposed [6, 7].

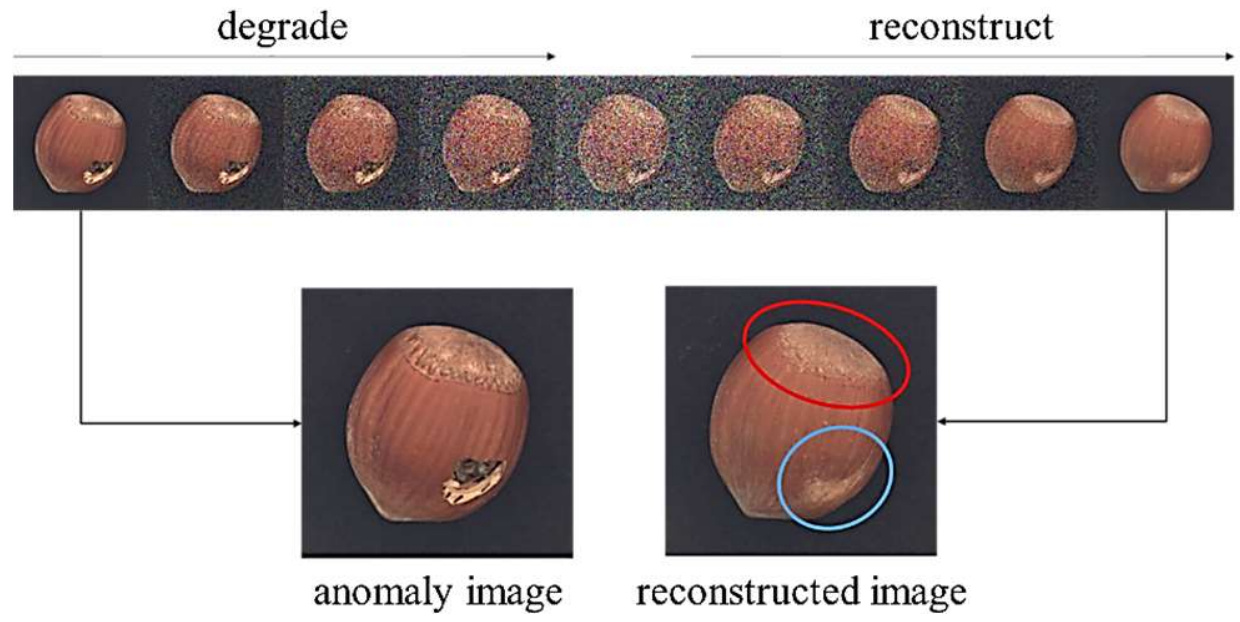

**Figure 1.** An example of the anomaly image and the reconstructed image. In addition to the anomaly (broken part) being converted to normal (blue), other normal regions with a lot of variety also have deformations (red).

To get the best result, it is important to create a good normal image to compare with, one that removes the anomaly while keeping the other information intact. This ensures that only anomaly regions are differentiated clearly without over- or under-detection. The process of generating a normal image from an anomaly through the diffusion model is composed of a forward process and a reverse process. Noise is first added to make it difficult to distinguish the anomaly from the normal, and then denoising is used to create a normal image that is close to the original [6, 7]. In this context, there are two challenges associated with generating a proper normalized image.

First, it is hard to determine the optimal noise level. The proper noise level is crucial because it is the mediator that converts an anomaly into something normal. If there is not enough noise, untrained anomalies may still be present, preventing the image from being restored to normal. On the other hand, if there is too much noise, the normal regions are degraded to the point where they cannot contain their original characteristics, resulting in many false positives. It is difficult to set the ideal noise level since it depends on the color, size, and shape of the unknown anomaly. Moreover, an image can even have multiple anomalies so it is impossible to determine one single value for all of them.

Second, false positives can be caused by the mode coverage property of the diffusion model. Diffusion models can produce a lot of variation even in normal regions and the reconstruction error can be high especially in regions with high contrast and fine detail.

As an example of the conventional method, Figure 1 illustrates the anomaly detection process on a hazelnut image. The broken part in the hazelnut is the target region to be detected as anomaly. The rightmost image is the reconstructed normal image to be compared to the original. Note that the anomaly region is converted to normal (blue), but another normal region with a fine detail also has been deformed (red), indicating the problem of false positives using the conventional method.

## 3. Method

To overcome the limitation of the conventional method, we propose an approach to analyze the trend of the reconstructions rather than just computing the reconstruction error for a particular noise level (Figure 2). As described above, in the reconstruction-based anomaly detection methods using diffusion, the noise level that degrades the image is an important parameter. By increasing the noise level and examining the trend of the reconstructed image as it changes, we obtain more information for a more reliable analysis, without relying on one critical parameter value. A gradually changing noise level will create a gradual change in the reconstructed image. The gradually changing noise level creates a gradual trend in the reconstruction image, where the anomaly region has a distinct trend from the normal region. If the noise level is insignificant, the anomaly will not be degraded enough, and it cannot be reconstructed to normal by the diffusion model. On the other hand, if the noise level is high enough, the anomaly will be sufficiently degraded to be indistinguishable from normal, and the diffusion model will be able to reconstruct it to normal. This means that as the noise level increases, the diffusion model is increasingly able to reconstruct the anomaly region similar to normal, and by analyzing each pixel to detect this gradual trend, anomaly region can be performed at the pixel level. Normal regions can also show some fluctuation in the reconstructed image due to different noise levels, but they are distinguished from anomalies by the absence of such large, gradual trends.

The existence of these trends in the reconstructed images is independent of the morphological characteristics of the anomalies, such as color, size, thickness, so the method solves the problem of properly setting the noise parameter for a range of unknown anomalies. It also mitigates the problem of false positives in normal regions because the fluctuations that occur in normal regions have a constant center and no significant trend, while anomalies change in one direction, slowly progressing from anomaly to normal.

In addition to analyzing the trend of the intensity, the proposed method also analyzes the trend of the model uncertainty at each pixel of the reconstructed image. Model uncertainty indicates how confidently the model predicts the clean data, and is higher on unfamiliar data not seen during training. For models trained only on normal data, the uncertainty will be higher for anomaly regions, which can be used as a cue for anomaly detection [9]. As the reconstructed image gets closer to normal with increasing noise level, the uncertainty tends to decrease, and we can identify this trend to further improve performance.

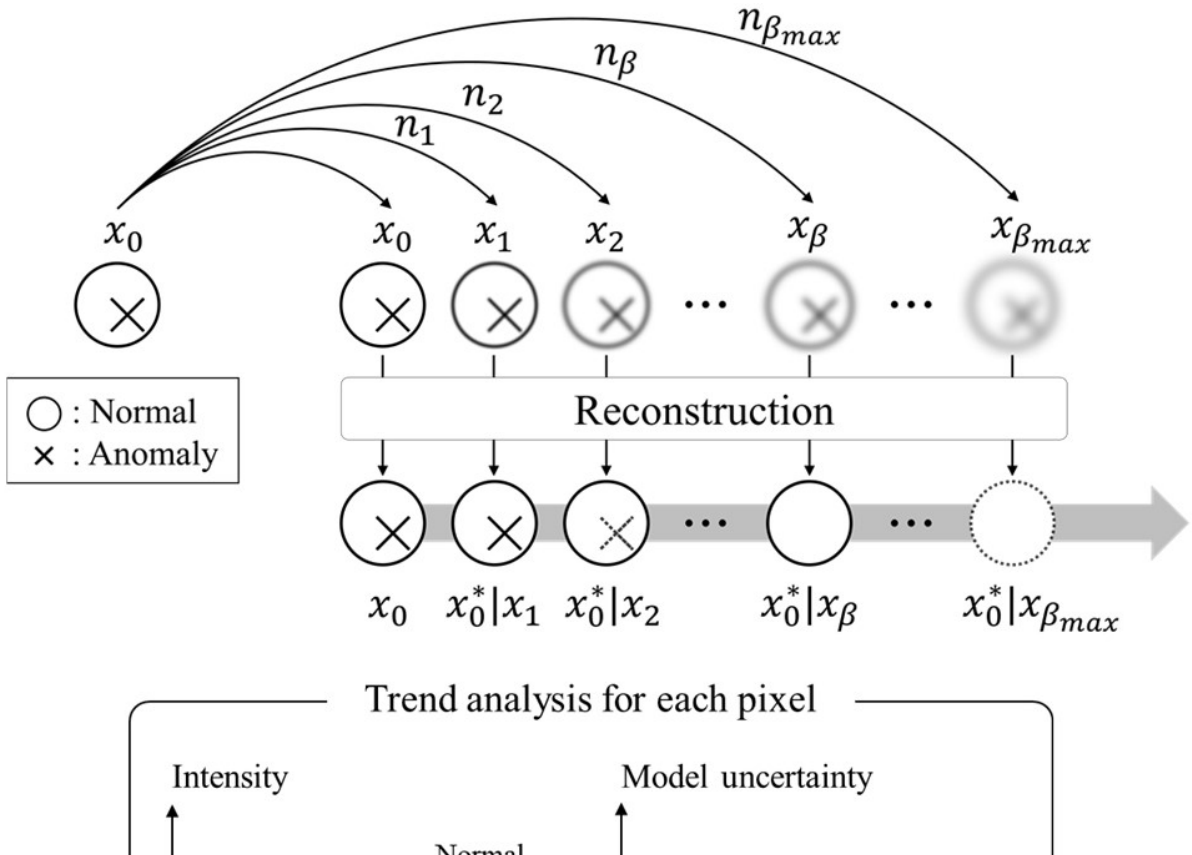

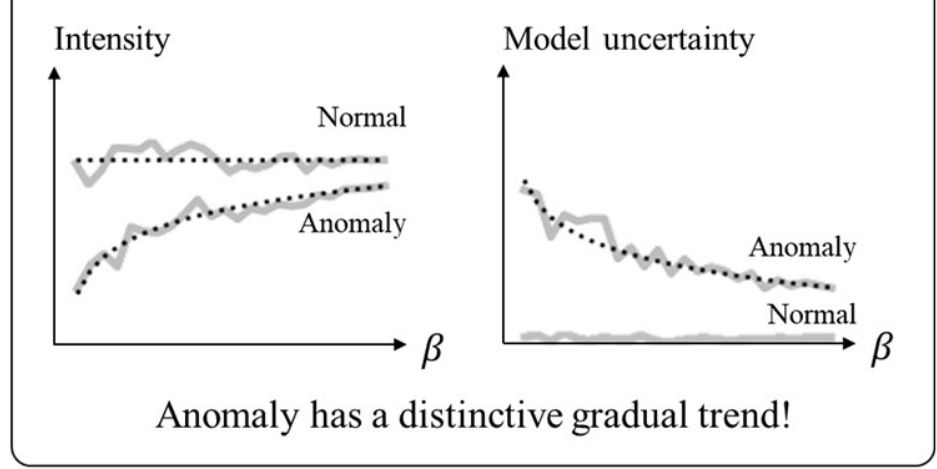

**Figure 2.** Illustration of the proposed method. The normal (○) and anomaly (×) regions are included in the original input image. The noise is added to generate degraded image, and then it is reconstructed back to normal by the diffusion model. As it degrades, the distinction between normal and anomaly dissipates, leading the anomaly to have a gradual trend toward normal in the reconstructed image, making it distinguishable from normal.

To this end, this study presents a novel method to quantify the uncertainty in diffusion models using model gradients. In the training of a neural network, a gradient conceptually means the change in the output image over each pixel change in the input image. Because the diffusion model is trained to counteract the noise in the input data, its predictions do not fluctuate much for small changes in values for familiar input data. On the other hand, for unfamiliar anomaly, the output becomes more susceptible to input changes. Since the model gradient reveals which regions of the image are susceptible to input changes, it can be used to detect anomaly. We calculated the gradient of the model output with respect to the model input. In this case, we need to get the anomaly trend of the reconstructed image, so we use reconstructed images with different noise levels as model inputs. The gradient of the denoising diffusion model is Jacobian, and it is computationally expensive to obtain the full matrix that describes the effect of every input pixel on the change in each output pixel. Therefore, we approximated it as the sum of gradients of output with respect to the input, which is the effect of the entire output image on each pixel in the input image. Given the assumption that only the adjacent parts will be affected by the input pixels, it gives us an idea of how sensitive the output pixels are to changes in the input pixels. This model gradient is easily computed as the calculation used for backpropagation.

We analyzed these two trends together. As the input anomaly image becomes increasingly noisy, the intensity and uncertainty in each pixel of the reconstruction image changes with a gradual and large trend. By combining these two independently calculated values, more reliable anomaly detection is possible.

Since these trends are caused by adding a small amount of noise, we have prior knowledge that the trend of interest is an extremely low-frequency component that changes slowly. To quantify the magnitude of the trend, we adopted the magnitude of the second Fourier coefficient which indicates a slow trend of the reconstructed signal changing with noise level. By removing all but the smallest frequency components, except for the DC component, we are able to effectively analyze these slowly changing trends. The resulting two trends were each normalized to have a value between 0 and 1 and combined by multiplication to get the final anomaly score (Figure 3).

**input**: input image: $x_0$, max noise level: $\beta_{max}$, diffusion model: $s$
**output**: anomaly score image: $score_anomaly$
**function** Anomaly-Detection ($x_0$, $\beta_{max}$, $s$)
    **for** $\beta$ from $0$ to $\beta_{max}$ **do**
        $n_\beta \leftarrow$ Gaussian noise image $\propto \beta$
        $x_\beta \leftarrow x_0 + n_\beta$
        $x_0^*|x_\beta \leftarrow$ prediction of $x_0$ from $x_\beta$ using $s$
        $u_\beta \leftarrow$ output gradient of $s$ at $x_0^*|x_\beta$
    **end**
    initialize three images: $trend_x$, $trend_u$, $score_anomaly$
    **for** each pixel $p$ **do**
        $trend_x_p \leftarrow$ 2nd component of FFT of $\{x_0, x_0^*|x_1, \ldots, x_0^*|x_{\beta_{max}}\}_p$
        $trend_u_p \leftarrow$ 2nd component of FFT of $\{u_0, u_1, \ldots, u_{\beta_{max}}\}_p$
    **end**
    $trend_x' \leftarrow$ normalized $trend_x$ to [0, 1]
    $trend_u' \leftarrow$ normalized $trend_u$ to [0, 1]
    $score_anomaly \leftarrow trend_x' \times trend_u'$
    **return** $score_anomaly$
**end**

**Figure 3.** Pseudocode for proposed method combining two trends for each pixel.

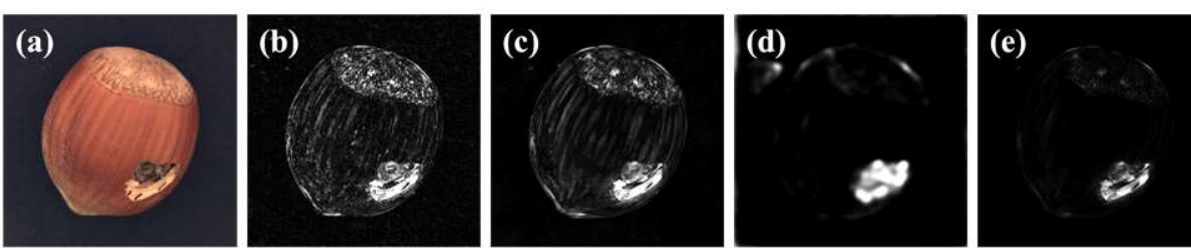

**Figure 4.** An example of anomaly detection showing (a) original image with anomaly, (b) conventional anomaly score obtained by analyzing reconstruction error, proposed anomaly score obtained by trending (c) intensity, (d) model uncertainty, and (e) the combination of the two.

## 4. Result

We evaluated our method using both the public MVTec dataset for industrial anomaly detection [10] and proprietary display inspection images, ensuring a comprehensive assessment across diverse application scenarios. While the proposed method can be combined with other reconstruction-based techniques, in this paper, we combined it with the based method, that calculates the reconstruction error with a single noise level using a diffusion model [5], to demonstrate the performance improvements. Figure 4 is an example of anomaly detection of hazelnuts (a), where a broken region corresponds to an anomaly.

We evaluated the proposed trend analysis-based anomaly detection technique against the existing method of comparing reconstruction-error (b). We set the two criteria to distinguish the anomaly: the trends of reconstruction error (e) and the model uncertainty (d), respectively. The result of the proposed method (e), which is the combination of the two trends, detects the anomaly cleanly and without false detection compared to the existing method (b). Figure 5 illustrates the evaluation results using proprietary display inspection images. To address confidentiality concerns, we utilized inspection images from previously released products. The evaluation focused on detecting various defect types in the wiring area, chosen to assess the method's effectiveness in challenging environments with detailed and sharp backgrounds. The results show that the proposed method has significantly fewer false detections due to the wiring background. 1000 normal images were used for training, while 20 unseen images for each defect type were used as the test set.

Tables 1 and 2 present the evaluation results for the MVTec dataset and the proprietary display inspection data, respectively. In many image anomaly detection cases, the normal and anomaly regions are often very unbalanced in area, and we are interested in detecting these small anomalies. In such scenarios, average precision (AP) can be important, since it is sensitive to rare event detection performance compared to the area under the receiver operating characteristic (AUROC) curve [11]. In this study, we added AP performance analysis in addition to AUROC, and combined the performance of all defect types to obtain mean AUROC (mAUROC) and mean AP (mAP). The proposed method performs well in all defect types and is particularly strong in mAP when uncertainty trend analysis is applied. The improvement in mAP is attributed to the increased sensitivity to small anomalies, the extent of which could be further controlled by adjusting the weight of uncertainty trend.

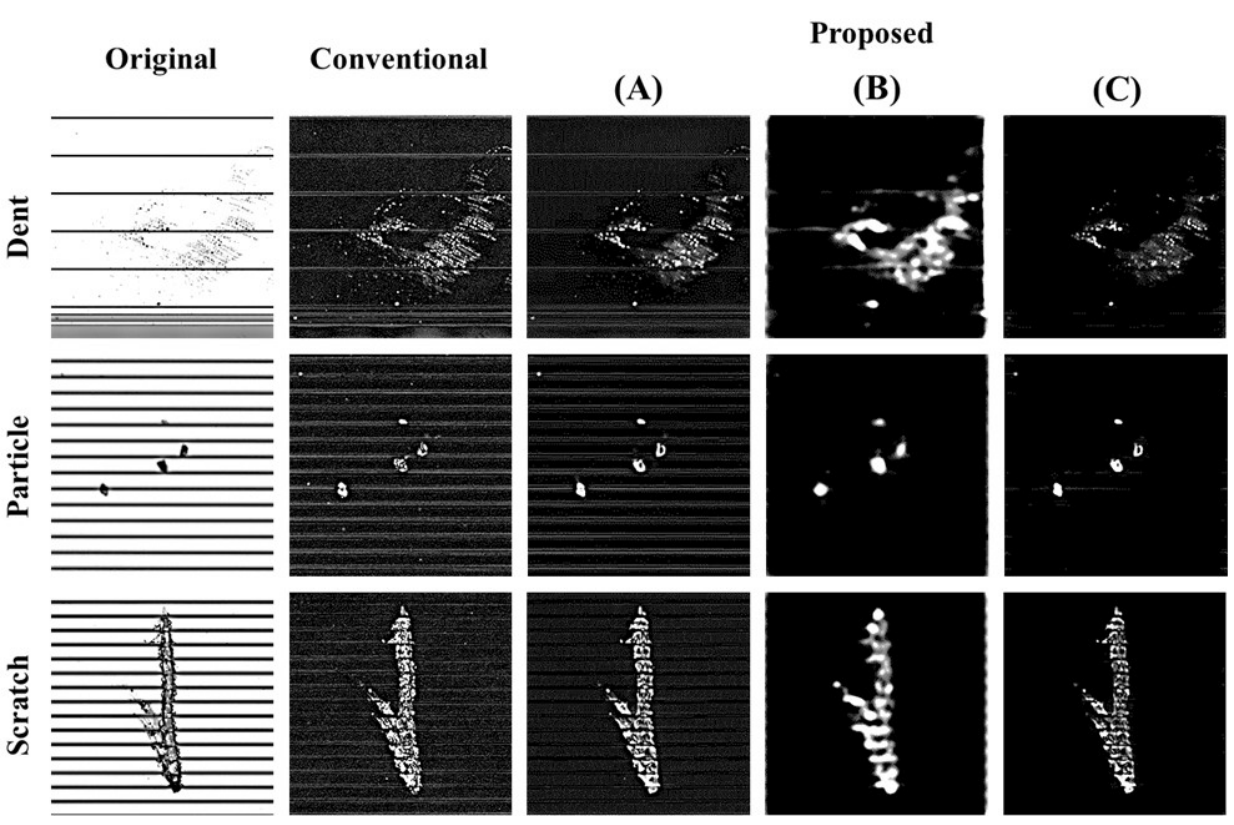

**Figure 5.** Example images for comparison of the conventional and proposed methods on anomaly detection tasks for various defects (dent, particle, and scratch) in display panel. Anomaly scores were obtained by trending the intensity (A) and model uncertainty (B) and combination of the two (C).

**Table 1.** The performance comparison between the conventional and proposed methods in anomaly detection tasks for various products. Anomaly scores were obtained by trending the intensity (A) and model uncertainty (B) and the combining the two (C).

| product | mAUROC (%) | | | | mAP (%) | | | |
|---|---|---|---|---|---|---|---|---|
| | before | proposed | | | before | proposed | | |
| | | (A) | (B) | (C) | | (A) | (B) | (C) |
| bottle | 87.0 | **88.7** | 74.3 | 70.0 | 38.7 | **41.3** | 33.5 | 35.5 |
| cable | 63.4 | 68.5 | **87.1** | 72.8 | 9.0 | 9.9 | **25.0** | 13.8 |
| capsule | 85.6 | 90.1 | **96.8** | 83.6 | 27.0 | 39.4 | **65.3** | 53.0 |
| carpet | 60.2 | 66.4 | **92.3** | 79.5 | 4.6 | 9.1 | **52.4** | 28.7 |
| grid | 82.8 | 92.6 | **99.4** | 95.9 | 24.1 | 38.8 | **60.8** | 59.0 |
| hazelnut | 94.1 | 96.9 | **97.9** | 94.3 | 41.1 | 53.5 | **73.3** | 71.6 |
| leather | 80.0 | 90.9 | **98.7** | 89.0 | 11.2 | 38.4 | **70.7** | 62.1 |
| metal nut | 83.1 | **86.7** | 59.0 | 77.5 | 32.1 | 33.5 | 32.9 | **44.7** |
| pill | 90.0 | **92.9** | 84.3 | 81.7 | **29.7** | 28.9 | 27.6 | 27.0 |
| screw | 88.0 | **95.9** | 95.2 | 91.6 | 20.6 | 31.1 | 17.5 | **44.0** |
| tansistor | 73.5 | 78.9 | **83.9** | 72.6 | 20.5 | 23.6 | **43.1** | 28.3 |
| tile | 61.8 | 64.3 | **81.5** | 65.1 | 15.4 | 19.5 | **47.1** | 26.4 |
| toothbrush | 93.4 | **95.0** | 93.4 | 91.7 | 16.8 | 20.9 | 18.4 | **22.8** |
| wood | 68.5 | 72.6 | **88.0** | 77.7 | 17.6 | 23.6 | **54.7** | 41.6 |
| zipper | 76.8 | **82.7** | 76.0 | 67.7 | 20.4 | **25.1** | 10.4 | 23.8 |
| average | 79.2 | 84.2 | **87.2** | 80.7 | 21.9 | 29.1 | **42.2** | 38.8 |

**Table 2.** The performance comparison between the conventional and proposed methods in anomaly detection tasks for various defects in display panel. Anomaly scores were obtained by trending the intensity (A) and model uncertainty (B) and the combination of the two (C).

| defects | mAUROC (%) | | | | mAP (%) | | | |
|---|---|---|---|---|---|---|---|---|
| | before | proposed | | | before | proposed | | |
| | | (A) | (B) | (C) | | (A) | (B) | (C) |
| dent | 85.8 | 92.2 | **92.6** | 92.4 | 30.8 | **33.8** | 16.8 | 32.1 |
| particle | 85.1 | 93.8 | **99.3** | 97.2 | 18.4 | 21.5 | **65.7** | 41.6 |
| scratch | 86.4 | 93.3 | **95.4** | 88.8 | 38.3 | **43.2** | 27.5 | 42.2 |
| average | 85.8 | 93.1 | **95.8** | 92.8 | 29.2 | 32.8 | 36.6 | **38.6** |

## 5. Conclusion

We proposed an unsupervised anomaly detection approach that can be combined to other diffusion-based reconstruction methods to provide a tradeoff between computational cost and performance. Moreover, the proposed method is flexible and compatible, allowing it to be used in conjunction with self-supervised or image-guided techniques, which can be useful in reconstruction-based anomaly detection. The proposed method requires minimal tuning and demonstrates robustness against false detection, making it highly valuable for rapid and efficient deployment in display manufacturing inspection processes characterized by diverse models, multiple production stages, and frequent introduction of new products.